\documentclass[conference]{IEEEtran}
\IEEEoverridecommandlockouts

\usepackage{cite}
\usepackage{amsmath,amssymb,amsfonts}
\usepackage{algorithmic}
\usepackage{graphicx}
\usepackage{textcomp}
\usepackage{xcolor}
\usepackage{pifont} 
\usepackage{multirow}
\usepackage{colortbl}
\usepackage[numbers]{natbib}         
\usepackage[colorlinks,
            linkcolor=blue,
            citecolor=blue,
            urlcolor=blue]{hyperref}
\def\BibTeX{{\rm B\kern-.05em{\sc i\kern-.025em b}\kern-.08em
    T\kern-.1667em\lower.7ex\hbox{E}\kern-.125emX}}
\usepackage[ruled,vlined]{algorithm2e}
\usepackage{diagbox}

\begin{document}

\title{Boosting Active Learning with Knowledge Transfer}
\author{
\IEEEauthorblockN{1\textsuperscript{st} Tianyang Wang}
\IEEEauthorblockA{\textit{University of Alabama at Birmingham}\\
Birmingham, United States\\
toseattle@siu.edu}
\and
\IEEEauthorblockN{2\textsuperscript{nd} Xi Xiao}
\IEEEauthorblockA{\textit{University of Alabama at Birmingham}\\
Birmingham, United States\\
xxiao@uab.edu}
\and
\IEEEauthorblockN{3\textsuperscript{rd} Gaofei Chen}
\IEEEauthorblockA{\textit{University of Alabama at Birmingham}\\
Birmingham, United States\\
gchen2@uab.edu}
\and
\IEEEauthorblockN{4\textsuperscript{th} Xiaoying Liao}
\IEEEauthorblockA{\textit{Johns Hopkins University}\\
Baltimore, United States\\
xliao13@jh.edu}
\and
\IEEEauthorblockN{5\textsuperscript{th} Guo Cheng}
\IEEEauthorblockA{\textit{Dalian University of Technology}\\
Dalian, China\\
2013012145@dlut.edu.cn}
\and
\IEEEauthorblockN{6\textsuperscript{th} Yingrui Ji\textsuperscript{\dag}}
\IEEEauthorblockA{\textit{Aerospace Information Research Institute, CAS} \& \\ University of Chinese Academy of Sciences}
Beijing, China\\
jiyingrui1996@gmail.com
\thanks{\dag\;Corresponding author.}}


\maketitle

\begin{abstract}
   Uncertainty estimation is at the core of Active Learning (AL). Most existing methods resort to complex auxiliary models and advanced training fashions to estimate uncertainty for unlabeled data. These models need special design and hence are difficult to train especially for domain tasks, such as Cryo-Electron Tomography (cryo-ET) classification in computational biology. To address this challenge, we propose a novel method using knowledge transfer to boost uncertainty estimation in AL. Specifically, we exploit the teacher-student mode where the teacher is the task model in AL and the student is an auxiliary model that learns from the teacher. We train the two models simultaneously in each AL cycle and adopt a certain distance between the model outputs to measure uncertainty for unlabeled data. The student model is task-agnostic and does not rely on special training fashions (e.g. adversarial), making our method suitable for various tasks. More importantly, we demonstrate that data uncertainty is not tied to concrete value of task loss but closely related to the upper-bound of task loss. We conduct extensive experiments to validate the proposed method on classical computer vision tasks and cryo-ET challenges. The results demonstrate its efficacy and efficiency.
\end{abstract}


\section{Introduction}
Supervised deep learning models achieve state-of-the-art performance but demand vast amounts of labeled data \cite{he2016deep,simonyan2014very}. This requirement is a significant bottleneck in specialized domains like Cryo-Electron Tomography (cryo-ET) classification, where unlabeled data is abundant but expert annotation is prohibitively expensive \cite{gubins2019shrec,chen2017convolutional}. Active learning (AL) addresses this challenge by iteratively selecting the most informative samples from an unlabeled pool for annotation, aiming to maximize model performance within a fixed labeling budget.

However, many existing AL methods suffer from practical drawbacks. Some rely on complex auxiliary components, such as variational autoencoders (VAEs) \cite{kingma2013auto}, generative adversarial networks (GANs) \cite{goodfellow2014generative,zhang2020state,sinha2019variational}, or task-specific loss prediction modules \cite{yoo2019learning}, which introduce significant tuning overhead and can be difficult to scale. Others are computationally inefficient, requiring either the solution to complex optimization problems \cite{sener2018active,kuo2018cost} or numerous forward passes through a network to approximate Bayesian uncertainty \cite{gal2016dropout,gal2017deep}. Furthermore, the concept of "uncertainty" itself is often loosely defined; for instance, we will show that equating it directly with predicted task loss, as some prior work does, is flawed.

To overcome these challenges, we propose a simple yet effective AL framework that leverages knowledge transfer to estimate data uncertainty. Our approach employs a teacher-student setup. The teacher model is the main task model, trained conventionally on the labeled set. Concurrently, a student model of the same architecture is trained with two objectives: the standard task loss and a knowledge transfer loss that encourages its feature representations to match those of the teacher. Our core hypothesis is that a sample is uncertain if the teacher and student models disagree on its prediction. We quantify this disagreement using the KL divergence between their posterior outputs and use this metric to select samples for annotation. This design elegantly avoids purpose-built auxiliary networks and complex training schemes, making it efficient and broadly applicable.

Theoretically, we prove that our selection strategy prioritizes data that results in a higher upper-bound of the task loss. We then establish a connection between this upper-bound and data uncertainty, providing a principled explanation for our method's effectiveness. We demonstrate the superiority of our approach through extensive experiments on image classification (CIFAR-10/100, SVHN, Caltech101, ImageNet), semantic segmentation (Cityscapes), and both simulated and real-world cryo-ET classification tasks, consistently outperforming state-of-the-art AL baselines.

We summarize our main contributions in three-fold.
\begin{enumerate}
    \item We exploit knowledge transfer to boost AL by proposing a new method to estimate uncertainty for unlabeled data. To our best knowledge, this is the first work to explore using knowledge transfer to estimate data uncertainty in AL.  
    \item We theoretically prove that our method selects data which leads to a higher upper-bound of task loss, making it possible to correlate data uncertainty with the upper-bound of task loss (section \ref{theory}). 
    \item We validate the proposed method on both computer vision and computational biology tasks. We also conduct comprehensive ablation studies to analyze the proposed method.  
\end{enumerate}

\section{Related Work}
\label{related}
Active learning strategies are typically categorized into three main families: uncertainty-based, diversity-based, and synthesis-based.

\noindent\textbf{Uncertainty-based.}
This is the most common approach in active learning, where the goal is to query samples for which the model is least certain. While traditional methods leveraged various uncertainty heuristics \cite{joshi2009multi, roth2006margin, tong2001support, xiao2025visualinstanceawareprompttuning, ji2025cibrcrossmodalinformationbottleneck, xiao2025visualvariationalautoencoderprompt}, the primary challenge in the deep learning era is estimating uncertainty effectively for modern neural networks. Recent works often address this by employing auxiliary models, such as Variational Auto-Encoders (VAEs) \cite{kingma2013auto} and adversarial training schemes \cite{sinha2019variational, zhang2020state}. Other approaches integrate deep models with classical optimization algorithms like k-centers to identify uncertain or core-set samples \cite{sener2018active, kuo2018cost}. Frameworks like transductive AL also leverage uncertainty estimation within a theoretically grounded context \cite{hubotter2024transductive}.

\noindent\textbf{Diversity-based.}
Diversity-based methods aim to select a batch of unlabeled samples that are representative of the overall data distribution, thereby ensuring the labeled pool is varied and non-redundant. These approaches often focus on data density or representation in a feature space \cite{hasan2015context, mac2014hierarchical, nguyen2004active, yang2015multi, zhang2025dpcoredynamicpromptcoreset}. Although historically treated as a separate paradigm, recent studies suggest a strong correlation between data diversity and model uncertainty \cite{loquercio2020general, demir2024}. This connection is explored in recent methods like MPTS, which samples along data manifolds to preserve structure and diversity \cite{ji2024deep}, and CAL, which selects representative samples across multiple domains \cite{hao2024composite}.

\noindent\textbf{Synthesis-based.}
A third paradigm involves synthesizing new, informative data points rather than selecting existing ones. These methods typically employ generative models like GANs \cite{goodfellow2014generative} or VAEs \cite{kingma2013auto} to create samples that are expected to be maximally beneficial for training the task model \cite{zhu2017generative, mahapatra2018efficient, mayer2020adversarial}. The utility of these synthesized samples is often evaluated using metrics like prediction entropy to guide the generation process.

\noindent\textbf{Discussion.}
Our work is an uncertainty-based method that contrasts with recent state-of-the-art approaches in its design simplicity and efficiency. Leading methods such as VAAL \cite{sinha2019variational} and State-Relabeling AL \cite{zhang2020state} depend on complex architectures with multiple, purpose-built modules (e.g., VAEs, discriminators) and adversarial training. Similarly, Learning Loss \cite{yoo2019learning} requires training a dedicated helper network just to predict task loss on unlabeled data.

While these methods have advanced the field, their reliance on highly customized components creates practical challenges. These auxiliary models often require careful tuning and can be difficult to adapt to new tasks or changes in input dimensions. In contrast, our proposed method is free of specialized modules and complex training paradigms. By leveraging a simple teacher-student framework, we provide a task-agnostic and efficient solution for estimating uncertainty.

\section{Methodology}
\label{method}
In this section, we present the proposed AL method. We firstly introduce its pipeline, including the motivation of using knowledge transfer in AL, training fashion of the teacher and the student model, uncertainty estimation and selection of unlabeled data. Then we analyze what type of data has higher uncertainty.



\subsection{Preliminaries}
\noindent
\textbf{Problem Formulation.}
Given an unlabeled data pool \{$X_U$\}, a labeling budget $K$ (percentage or fixed size), and an initially empty labeled pool \{$X_L$, $Y_L$\}, active learning (AL) aims to select $K$ samples from \{$X_U$\} for annotation by human oracles. The annotated samples \{$X_S$, $Y_S$\} are then added to the labeled pool: \{$X_L$, $Y_L$\} $\leftarrow$ \{$X_L$, $Y_L$\} $+$ \{$X_S$, $Y_S$\}, while removing them from the unlabeled pool: \{$X_U$\} $\leftarrow$ \{$X_U$\} $-$ \{$X_S$\}. The task model is trained on \{$X_L$, $Y_L$\} using a supervised loss (e.g., cross-entropy for classification). These steps are repeated for multiple AL cycles until the labeling budget is met. We use $T$ and $T(\cdot)$ for the task model and its forward operation, and $S$ and $S(\cdot)$ for the student model and its forward operation.

\noindent
\textbf{Knowledge Transfer.} In knowledge transfer, a student model can learn from a teacher to achieve similar performance \cite{hinton2015distilling,zagoruyko2016paying}. For this to be effective, the student should have the same number of modules as the teacher, allowing knowledge to be transferred between corresponding model parts \cite{zagoruyko2016paying}. While the models are trained to agree, we conjecture that if the teacher and student have different opinions on an unlabeled sample, that sample has higher uncertainty. Therefore, the distance between their outputs can be used to estimate data uncertainty.

\subsection{Method Pipeline}
\label{pipeline}

\noindent
\textbf{Models and Training.}
Our framework uses a teacher-student structure for knowledge transfer in AL. Following \cite{zagoruyko2016paying}, we set the student to have the same number of modules (layer blocks yielding different resolutions) as the teacher. For example, ResNet-18~\cite{he2016deep} and VGG-16~\cite{simonyan2014very} serve as teacher models, with ResNet-10~\cite{he2016deep,zagoruyko2016paying} and VGG-13~\cite{simonyan2014very} as corresponding students, since their module counts match.

In each AL cycle, both teacher and student are trained together. The teacher uses only task loss, while the student is optimized with both task loss and a knowledge transfer loss. The total loss is $L = L_{task} + \lambda L_{trans}$, where $\lambda$ balances the two terms (set to $100$ in our experiments). Following~\cite{hinton2015distilling,zagoruyko2016paying}, we compute $L_{trans}$ over feature space, which is more effective than probability space losses. We adopt the attention-transfer loss~\cite{zagoruyko2016paying}, which outperforms MSE and L1 losses (see Fig.~\ref{fig:dist} \textbf{Left}). Specifically,
\vspace{-0.2cm}
\begin{equation}
  L_{trans} = \sum_{l=1}^N\left\| \frac{V_S^l}{\|V_S^l\|_2} - \frac{V_T^l}{\|V_T^l\|_2} \right\|_2,
  \label{eq2}
\end{equation}
where $V_S^l$ and $V_T^l$ are vectorized attention maps at the $l$-th module, $N$ is the number of module pairs, and the attention map is computed as $\sum_{i=1}^C |A_i|^2$, with $A_i$ as the feature activation and $C$ the channel number.

\noindent
\textbf{Uncertainty Estimation.}
After training models $T$ and $S$ in each AL cycle, we estimate data uncertainty using both models. For each unlabeled sample $X_U$, we define uncertainty as
\vspace{-0.1cm}
\begin{equation}
\small
\label{eq3}
  uncertainty(X_U) = f_D(T(X_U), S(X_U)),
\end{equation}
where $f_D$ measures the distance between the outputs of $T$ and $S$. For classification, we find that using KL divergence on the softmax outputs gives the best results (see Fig.~\ref{fig:dist} \textbf{Right}), so
\vspace{-0.1cm}
\begin{equation}
\small
\label{eq4}
  uncertainty(X_U) = KL_{div}(s(T(X_U)), s(S(X_U))),
\end{equation}
where $s(\cdot)$ is the softmax function. For semantic segmentation, we use MSE on the probability maps:
\vspace{-0.1cm}
\begin{equation}
\small
\label{eq5}
  uncertainty(X_U) = MSE(s(T(X_U)), s(S(X_U))).
\end{equation}

Note that we use different metrics for student training and for uncertainty estimation: attention-transfer loss over feature space for training, and KL divergence or MSE over probability space for uncertainty estimation. This combination achieves the best empirical performance, as shown in the ablation study.

\noindent
\textbf{Data Selection.}
Prior work~\cite{sener2018active,settles2012active} shows that selecting from the entire unlabeled pool is often suboptimal. Following~\cite{beluch2018power,sener2018active,settles2012active,yoo2019learning}, we first sample a subset $\{R_U\} \subset \{X_U\}$ and perform data selection within this subset. For fair comparison, this strategy is used for all methods in Section~\ref{experiment}. We set $len(\{R_U\}) \approx 10 \times M$, where $M$ is the number of samples to select per AL cycle. We estimate the uncertainty of each sample in $\{R_U\}$ using eq.~\ref{eq4} or \ref{eq5} and select the top $M$ most uncertain samples for annotation.

\begin{figure}[t]
    \centering
    \includegraphics[width=0.8\linewidth]{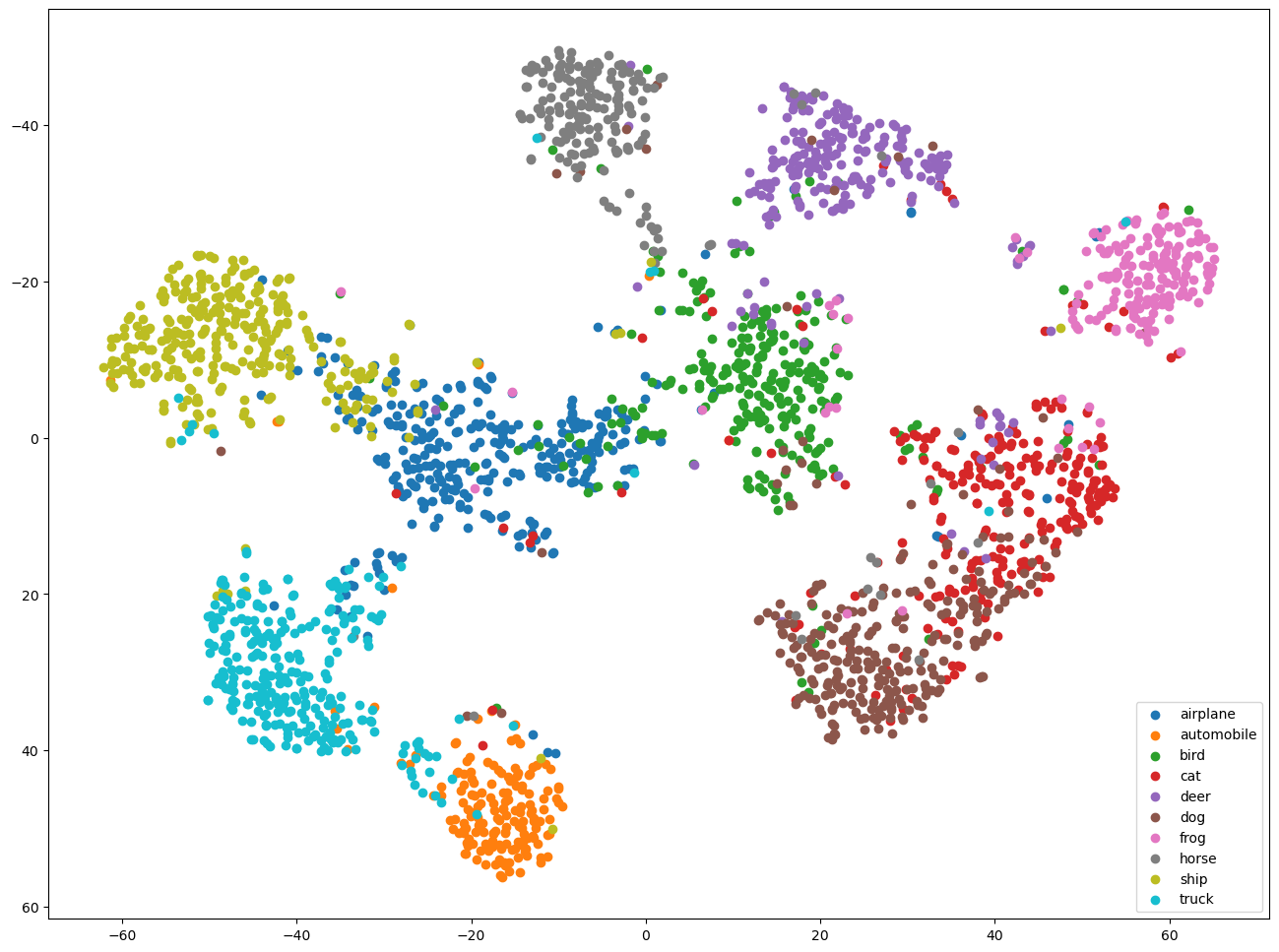}
    \includegraphics[width=0.8\linewidth]{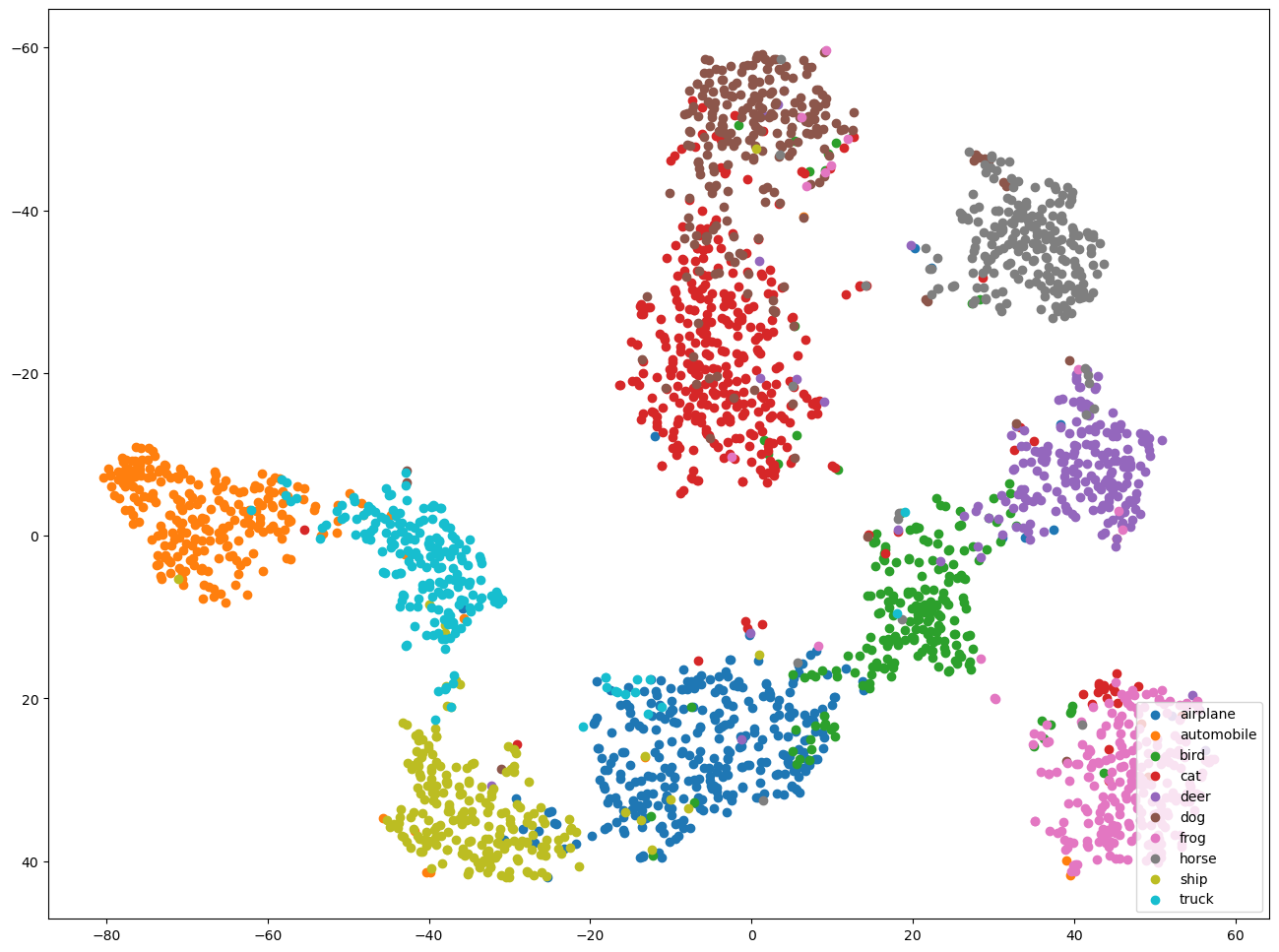}
    \caption{t-SNE visualization of the 2500 data samples selected by LL4AL~\cite{yoo2019learning} (top) and our method (bottom) from Cifar10~\cite{krizhevsky2009learning} using ResNet-18~\cite{he2016deep}.}
    \label{fig:tsne}
    \vspace{-0.3cm}
\end{figure}

\subsection{Data Uncertainty and Task Loss}
\label{theory}

Here, we connect data uncertainty with the upper bound of task loss. First, we show that our method selects data leading to a higher upper-bound of $L_{task}$, then demonstrate that data uncertainty is closely related to this upper-bound.

Let $x$ be labeled training data, $w$ and $\theta$ be the weights of the teacher $T$ and student $S$, and $g(\cdot)$ denote the network operations. The distance between $T$ and $S$ is
\vspace{-0.2cm}
\begin{equation}
\small
\label{eq6}
  D = \|g(w_{c}x) - f(\theta_{c} x)\| ,
\end{equation}
where $c$ is the current AL cycle.

During training at the $c$-th cycle, $S$ minimizes $D$ to match features with $T$ (see eq.~\ref{eq2}). For data selection, we choose samples with larger $D$.

In the $(c+1)$-th cycle, the task loss is $L_{task} = \|g(w_{c}x_s) - y\|$, where $y$ is the label for $x_s$. We can write
\vspace{-0.1cm}
\begin{equation}
\small
\label{eq7}
\begin{split}
  L_{task} &= \|g(w_{c}x_s) - y\| \\
  & \leq \|g(w_{c}x_s) - f(\theta_{c} x_s)\| + \|f(\theta_{c} x_s) - y\| \\
  & = D + \|f(\theta_{c} x_s) - y\| .
\end{split}
\end{equation}
A larger $D$ implies the student is farther from the teacher, and likely farther from the true label $y$, so $L_{task}$ is upper-bounded by a higher value. Thus, our method tends to select samples that contribute more to the upper bound of the task loss.

Next, we show that our selected data has higher uncertainty. We evaluate the task model trained on data selected by our method and compare it with LL4AL~\cite{yoo2019learning}, which measures uncertainty using task loss. First, we randomly select 17,500 samples from the Cifar10~\cite{krizhevsky2009learning} training set to train the task model. Using the trained model, we then select 2,500 additional unlabeled samples with both our method and LL4AL from the remaining 32,500 samples. The selected data has not been seen during initial training.

We use t-SNE~\cite{van2008visualizing} to visualize the selected samples (Fig.~\ref{fig:tsne}). Our method results in more correctly classified samples. Specifically, the classification accuracy on the selected data is 87.76\% for our method and 84.4\% for LL4AL. This shows our selected data has lower $L_{task}$.

However, when we retrain the model with both the initial 17,500 samples and the 2,500 selected samples, our method achieves 93.03\% accuracy on the Cifar10 test set, while LL4AL yields 91.3\%. This indicates our selected data is more informative and uncertain, since it leads to better performance in the AL setting. These results support that data uncertainty is not strictly determined by task loss, but is related to the upper-bound of the task loss.

\subsection{Discussion}
\label{discussion}

From another perspective, our method can be viewed through the lens of noise robustness. Feeding data to the student model is similar to introducing noise to the teacher. The student learning from the teacher acts as a denoising step. If both models give similar outputs, the data is less uncertain and more robust; if their outputs differ, it suggests higher uncertainty due to the impact of noise.

\section{Experiments}
\label{experiment}

We conduct extensive experiments to evaluate the effectiveness and efficiency of our method. First, we validate performance on standard image classification and semantic segmentation benchmarks. We then assess our approach on both simulated and real cryo-ET datasets to demonstrate its suitability for specialized domain tasks. Finally, we present ablation studies for detailed analysis. All experiments follow established protocols, and we reproduce baseline results using the original settings. The labeling budget ranges from 10\% to 40\% (or 30\%) in steps of 5\%, yielding 7 (or 5) AL cycles. In the first cycle, 10\% of unlabeled data is randomly selected and annotated as the initial training set for all methods. We compare our method with state-of-the-art AL baselines: MC-Dropout~\cite{gal2016dropout}, QBC~\cite{kuo2018cost}, Core-Set~\cite{sener2018active}, VAAL~\cite{sinha2019variational}, LL4AL~\cite{yoo2019learning}, SRAAL~\cite{zhang2020state}, and random selection. Results are averaged over 3 runs (2 runs for ImageNet). For clarity, we recommend zooming in on all figures.
 

\begin{figure*}[t]
    \centering
    \includegraphics[width=0.32\linewidth]{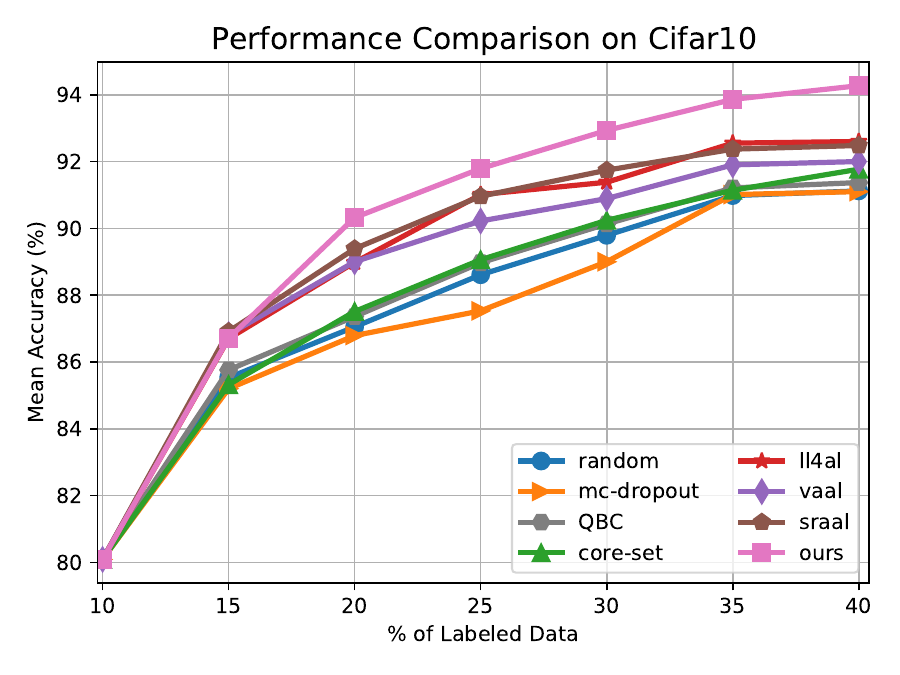}
    \includegraphics[width=0.32\linewidth]{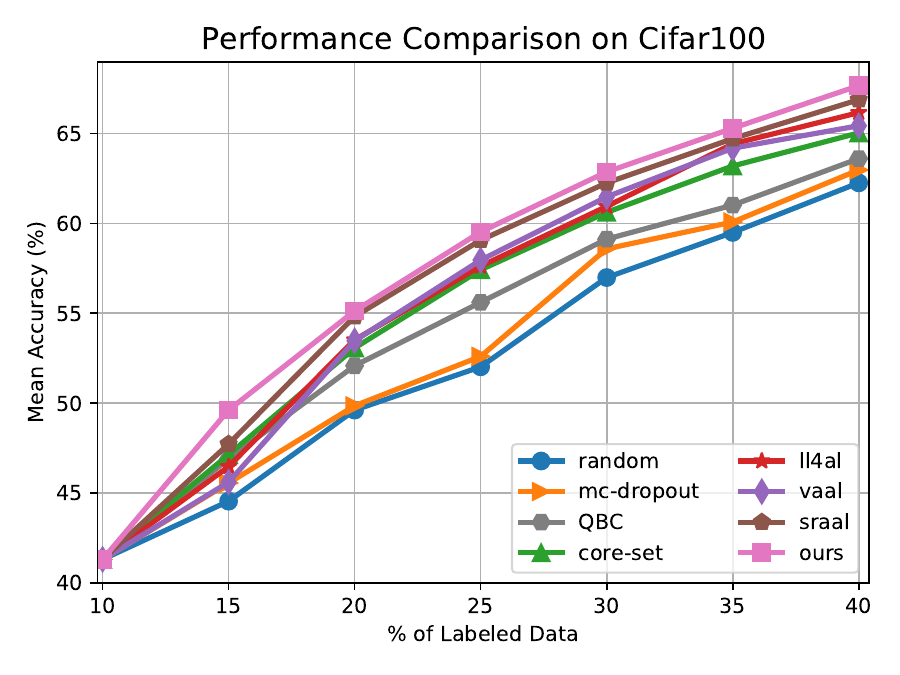}
    \includegraphics[width=0.32\linewidth]{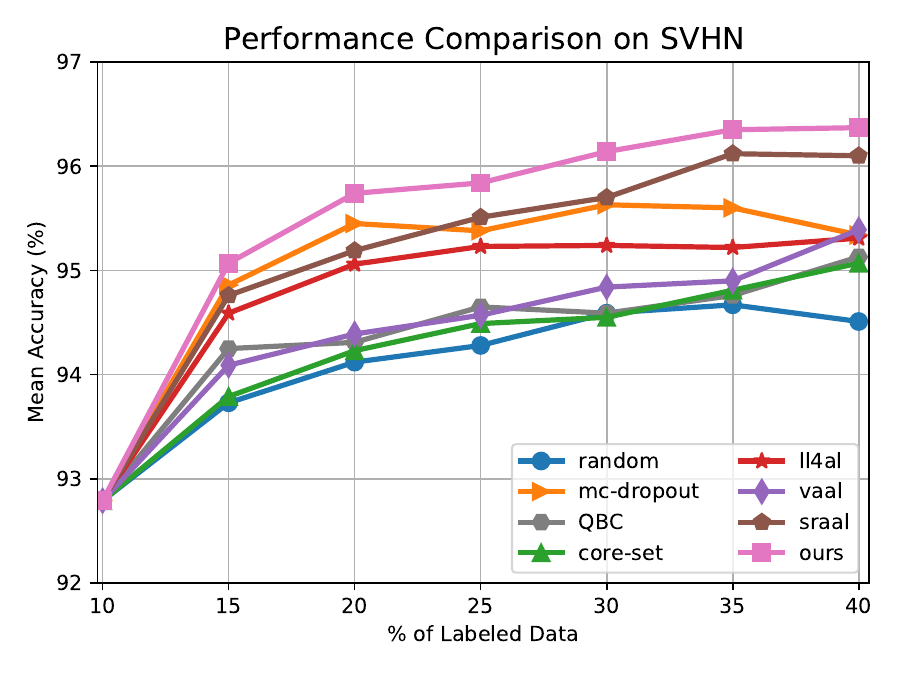}
    \caption{Performance comparison of the AL methods on Cifar10 (\textbf{Left}), Cifar100 (\textbf{Middle}), and SVHN (\textbf{Right}), respectively.}

    \label{fig:classification1}
\end{figure*}

\begin{figure*}[t]
    \centering
    \includegraphics[width=0.32\linewidth]{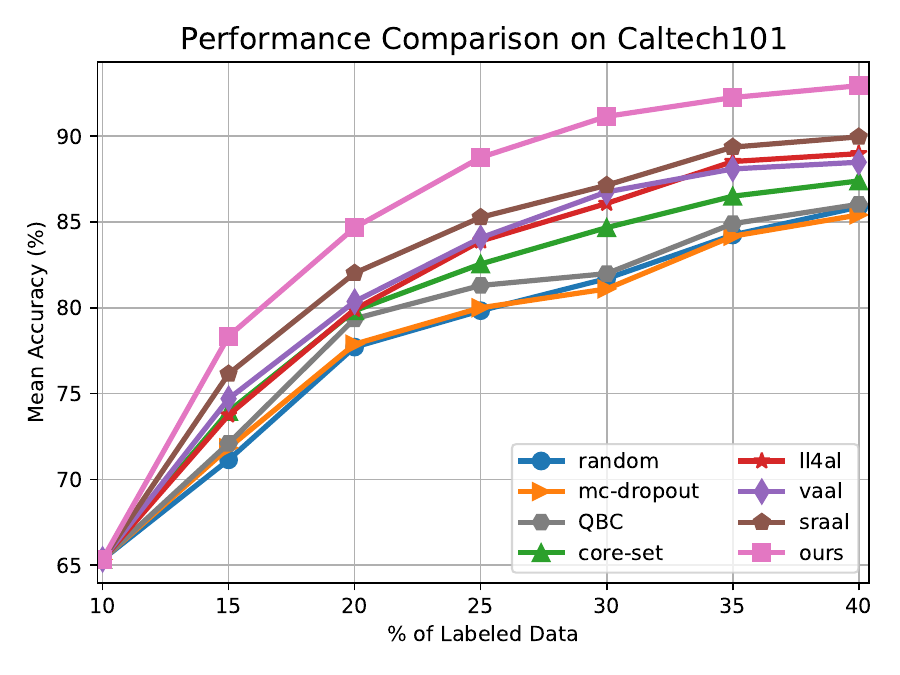}
    \includegraphics[width=0.32\linewidth]{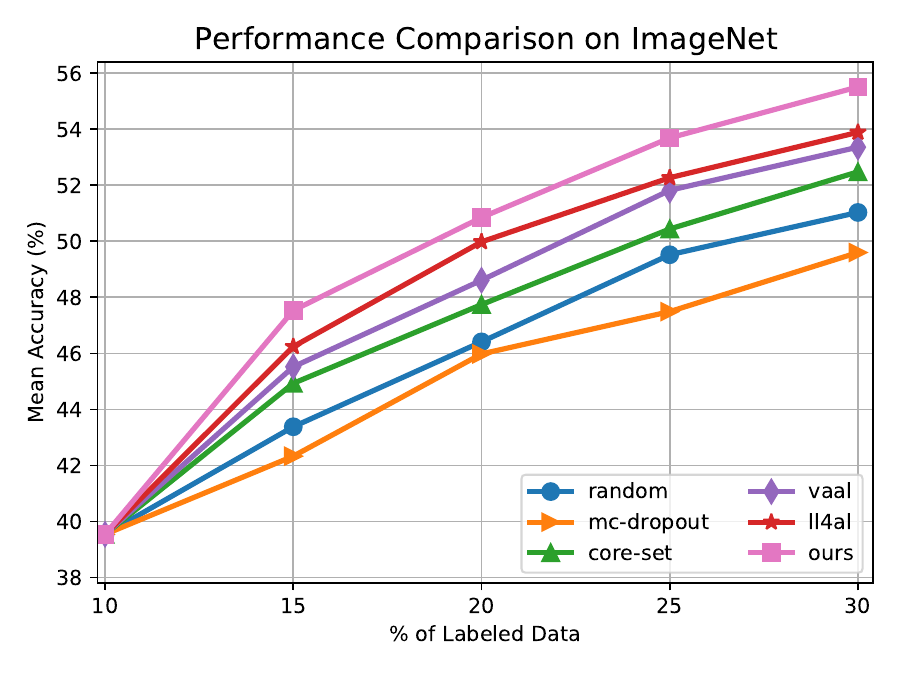}
    \includegraphics[width=0.32\linewidth]{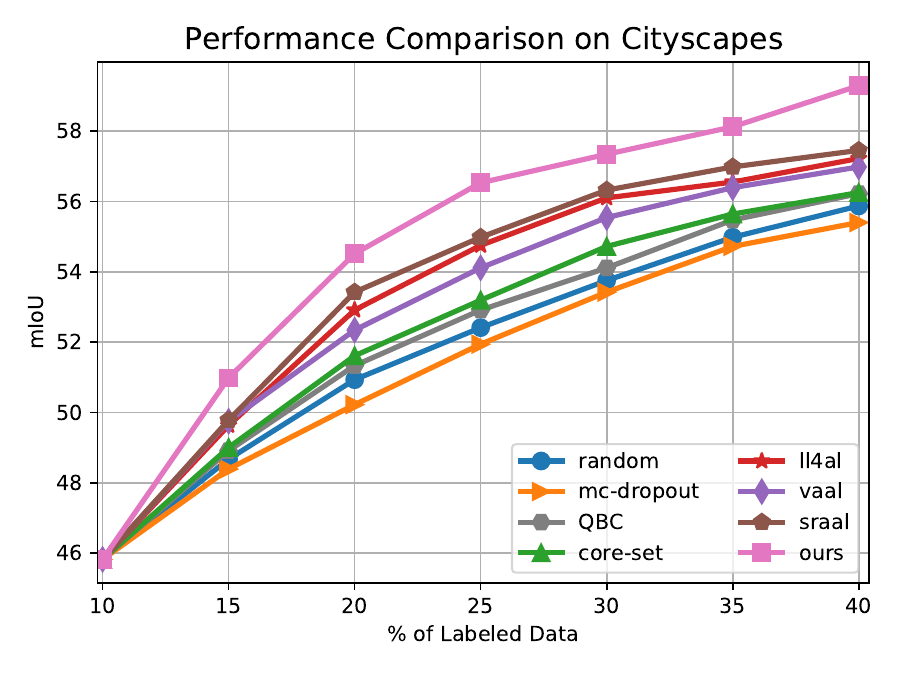}
    \caption{Performance comparison on Caltech101 (\textbf{Left}), ImageNet (\textbf{Middle}), and Cityscapes (\textbf{Right}). The first two are classification tasks while the last one is a \textbf{segmentation task.}}
    \label{fig:classification2}
    \vspace{-0.3cm}
\end{figure*}


\subsection{Image Classification}
\label{imgclassify}

\noindent
\textbf{Datasets.} We evaluate our method on five benchmark classification datasets: Cifar10~\cite{krizhevsky2009learning}, Cifar100~\cite{krizhevsky2009learning}, SVHN~\cite{svhn}, Caltech101, and ImageNet~\cite{deng2009imagenet}. Cifar10 and Cifar100 contain 50,000 training and 10,000 test images, with 10 and 100 classes, respectively. SVHN has 73,257 training and 26,032 testing samples (we exclude extra training data for fairness). Caltech101 includes 8,677 images, with class sizes ranging from 40 to 800, and is imbalanced. ImageNet is a large-scale dataset with about 1.28 million training images over 1,000 classes, and a 50,000-image validation set.

\noindent
\textbf{Model Selection.} For Cifar and SVHN, we use a standard ResNet-18~\cite{he2016deep} variant suited for $32\times32$ images~\cite{zhang2020state}. For Caltech101, we use the original ResNet-18. To ensure a fair comparison, all methods use the same architecture for each dataset. For example, in Cifar10, we replace VGG-16 with ResNet-18 for methods like Core-Set~\cite{sener2018active} and VAAL~\cite{sinha2019variational}. To show model-independence, VGG-16 is used for all methods on ImageNet. The student model is a shallower version of the teacher: ResNet-10~\cite{zagoruyko2016paying} for ResNet-18, and VGG-13 for VGG-16.

\noindent
\textbf{Training Details.} 
In the two Cifar and SVHN experiments, we train the task model and the student for 200 epochs with a batch size of 128 and an initial learning rate of 0.1, which is decayed by a factor of 0.1 at the $160^{th}$ epoch. For Caltech101, we train the two models for 50 epochs. The batch size is set to 64 due to the larger image size. The initial learning rate is set to 0.01 and decayed by a factor of 0.1 at the $40^{th}$ epoch. We adopt the SGD optimizer \cite{bottou2010large} for all these experiments and set the momentum and weight decay rate to 0.9 and $5\times10^{-4}$, respectively. For ImageNet, we train the two models for 100 epochs with a batch size of 64. We use the Adam optimizer \cite{kingma2014adam} with a learning rate of 0.1 through the entire training process.

\noindent
\textbf{Results and Analysis.}
As shown in Fig.~\ref{fig:classification1} and~\ref{fig:classification2}, our method consistently outperforms all baselines across datasets. For every labeling budget, our approach achieves higher accuracy, which is valuable for practical AL scenarios where the budget may vary. 

On Cifar10, our method reaches 91.79\% accuracy with 12.5K labeled samples, while SRAAL and VAAL require 2.5K and 5K more samples, respectively, for similar results. Our approach also shows clear improvements on SVHN and Caltech101, highlighting its robustness to class imbalance. The superior results on ImageNet further verify our method’s scalability to large-scale tasks. Notably, on Cifar10, our method achieves 94.27\% accuracy, surpassing the result (93.6\%) obtained by training on the full dataset. This supports findings in~\cite{koh2017understanding} that some training samples may negatively impact deep model learning.

\subsection{Semantic Segmentation}
\label{seg-sec}
We evaluate our method on the semantic segmentation task to demonstrate its task-agnostic nature. 

\noindent
\textbf{Dataset.}
We use the Cityscapes dataset \cite{cityscapes}, which is large-scale and consists of images of street scenes from 50 cities. The images are taken under various weather conditions in different seasons, leading to a challenging task. For fair comparison, we only adopt the standard training and validation data. Following the practice in \cite{yu2017dilated}, we crop the images to a dimension of $688\times688$ and choose 19 categories for pixel-level classification. 

\noindent
\textbf{Model Selection.}
We adopt a dilated residual network, namely DRN-D-22 \cite{yu2017dilated}, as the task model. We use DRN-D-14 as the student. Both networks have 8 layer modules.

\noindent
\textbf{Training details.}
We train the teacher and student model for 50 epochs with the Adam optimizer \cite{kingma2014adam}. The initial learning rate is set to $5\times10^{-4}$ and decayed at the $40^{th}$ epoch by a factor of 0.1.

\noindent
\textbf{Results and Analysis.} The mIoU (mean Intersection-over-Union) is employed to measure the performance. As illustrated in Fig. \ref{fig:classification2} \textbf{Right}, our method yields solid better results than the others. 
Since the Cityscapes is highy imbalanced, this experiment once again demonstrates the superiority of our method on imbalanced datasets. More importantly, from classification to segmentation, the student model in our method is free of complex design,
whereas VAAL \cite{sinha2019variational} and SRAAL \cite{zhang2020state} must redesign their auxiliary models (e.g. VAE) to adapt to different input format. This makes our method convenient to use in various tasks.

\subsection{Cryo-ET Challenges}
\label{cryo-sec}
Active learning is particularly valuable in domain tasks~\cite{kuo2018cost, gorriz2017cost, konyushkova2017learning}, where annotation costs are high. We evaluate our method on cryo-ET classification tasks.

\noindent
\textbf{Datasets.} We use two cryo-ET datasets: a simulated set (50c-snr005) with SNR 0.05~\cite{liu2020efficient}, containing 24,000 training and 1,000 test samples evenly spread across 50 classes; and a real dataset (10c-real)~\cite{noble2018routine,guo2018situ} from medical practice, with 4,318 training and 1,080 testing samples, and notable class imbalance (320–876 samples per class for training, 80–219 for testing).

\noindent
\textbf{Model Selection.} Cryo-ET data consists of $32\times32\times32$ voxel grids, so we use a 3D ResNet-18 as the task model for all methods. Competing approaches like VAAL~\cite{sinha2019variational}, SRAAL~\cite{zhang2020state}, and LL4AL~\cite{yoo2019learning} require task-specific auxiliary model redesign for 3D input, whereas our method only replaces the student with a 3D ResNet-10, with minimal modification.

\noindent
\textbf{Training Details for Simulated Data.}
Both task and student models are trained for 100 epochs with initial learning rate 0.1 (decayed by 0.1 at epoch 80), using SGD~\cite{bottou2010large} with momentum 0.9 and weight decay $5\times10^{-4}$.

\begin{figure}[t]
    \centering
    \includegraphics[width=0.48\linewidth]{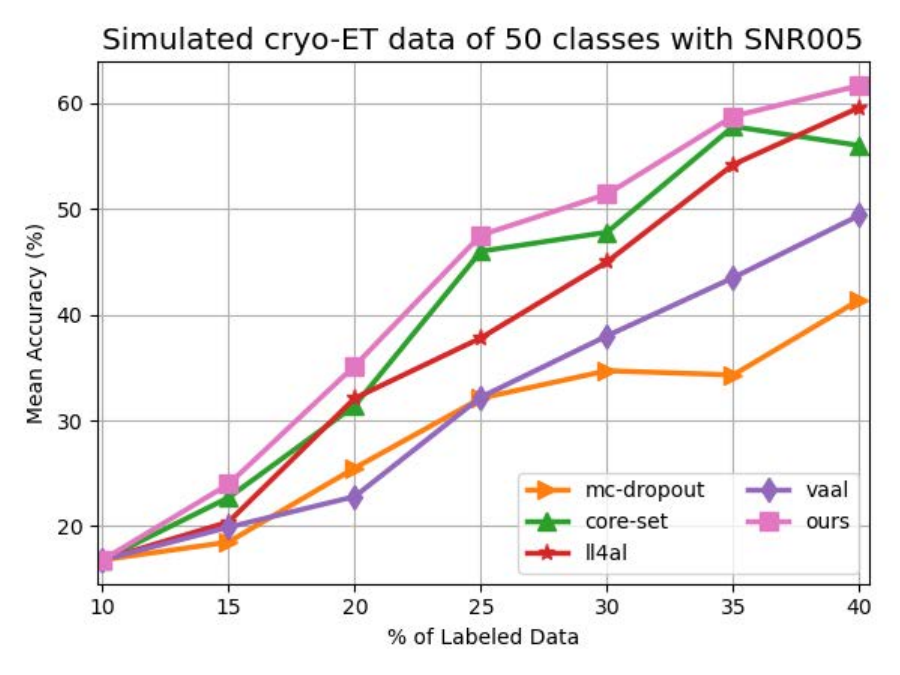}
    \includegraphics[width=0.48\linewidth]{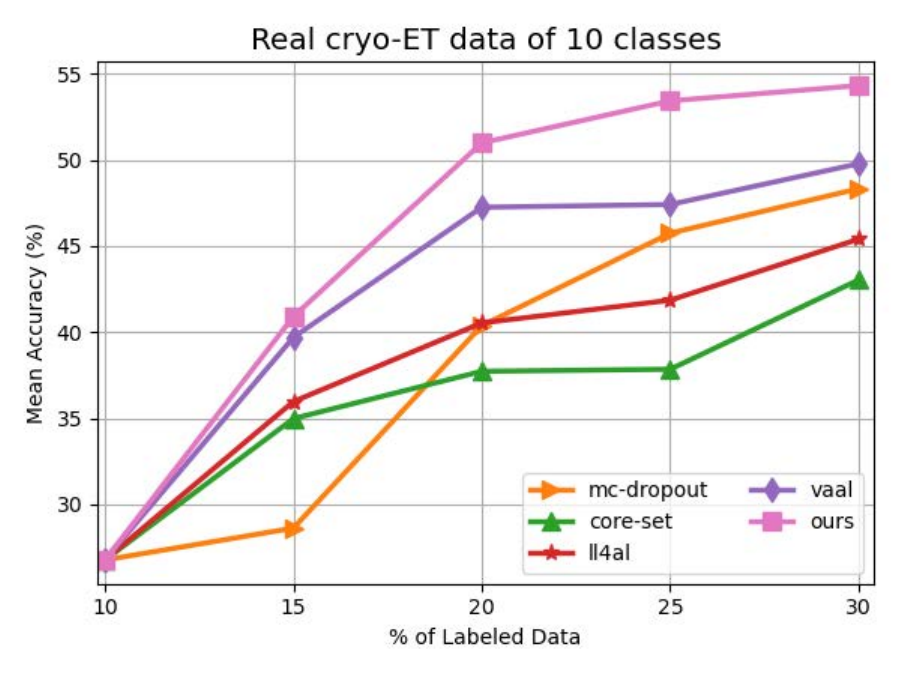}
    \caption{Performance comparison on simulated (\textbf{Top}) and real cryo-ET data (\textbf{Bottom}). 
    }
    \label{fig:cryoET}
    \vspace{-0.3cm}
\end{figure}

\noindent
\textbf{Training Details for Real Data.}
Due to the much fewer samples in the real cryo-ET dataset, we make the labeling budget vary from 10\% to 30\% with an incremental size of 5\%, corresponding to 5 AL cycles. We train the two models for 5 epochs within each cycle to avoid over-fitting. We also use the SGD with the same settings as in the simulated data, except that the learning rate of 0.1 will not be decayed during training.

\noindent
\textbf{Results and Analysis.}
We illustrate the results in Fig. \ref{fig:cryoET}. As can be seen, our method yields constantly better results on the two datasets, demonstrating its reliability. Moreover, our method outperforms the others by a large margin on the simulated data, which contains high level of noise (i.e. SNR005), demonstrating its robustness to noise.  
For the real data, the competence of our method is not impaired by the small number of training samples, indicating that the task model trained by our selected data can generalize better on testing data. 

\subsection{Ablation Study}
\begin{figure}[htbp]
\vspace{-0.2cm}
    \centering
    \includegraphics[width=0.49\linewidth]{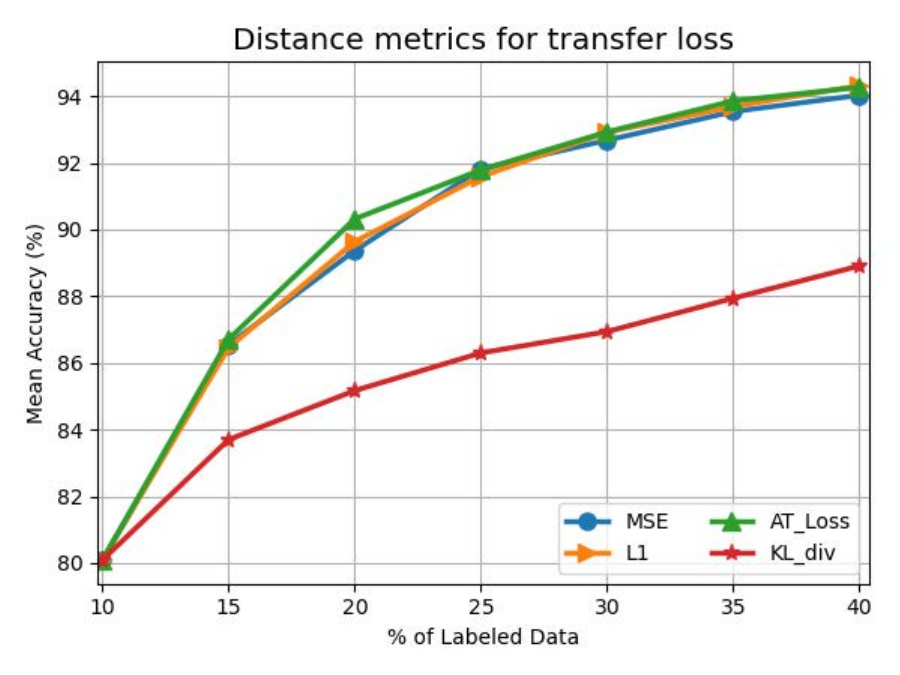}
    \includegraphics[width=0.49\linewidth]{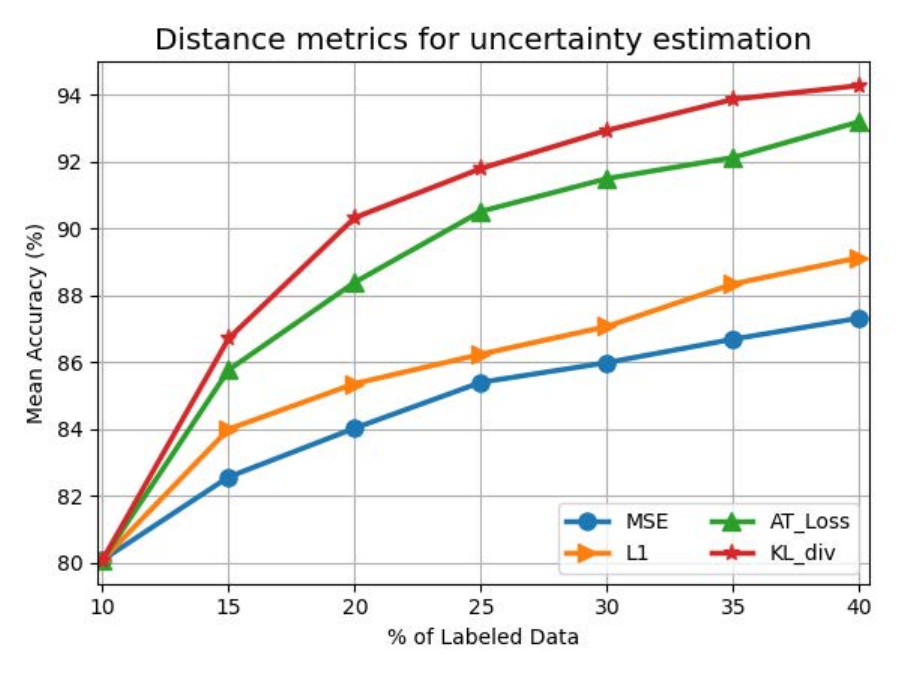}
    \caption{Distance metrics for the transfer loss (\textbf{Left}) and for the uncertainty estimation (\textbf{Right}). These experiments are conducted on Cifar10 \cite{krizhevsky2009learning}.
    }
    \label{fig:dist}
    \vspace{-0.2cm}
\end{figure}

\noindent
\textbf{Distance Metric for Transfer Loss.}
As shown in eq.~\ref{eq2}, we use the attention-transfer loss~\cite{zagoruyko2016paying} for student training. We also compare different distance metrics, including MSE, L1 (both in feature space), and KL divergence over posteriors. As illustrated in Fig.~\ref{fig:dist}~\textbf{Left}, attention-transfer, MSE, and L1 achieve similar performance, while KL divergence on posteriors performs significantly worse. Therefore, we recommend using a feature space distance metric for transfer loss, and adopt attention-transfer loss in our experiments due to its slightly better results.

\noindent
\textbf{Distance Metric for Uncertainty Estimation.}
For uncertainty estimation, we find that KL divergence over output posteriors significantly outperforms distance metrics in feature space, such as MSE, L1, or attention-transfer loss (see Fig.~\ref{fig:dist}~\textbf{Right}). Thus, we use KL divergence on posteriors for classification. For segmentation, although we use MSE (eq.~\ref{eq5}), it is still computed over the probability maps (posteriors) at each pixel, not in feature space.

\begin{figure}[htbp]
    \centering
    \includegraphics[width=0.8\linewidth]{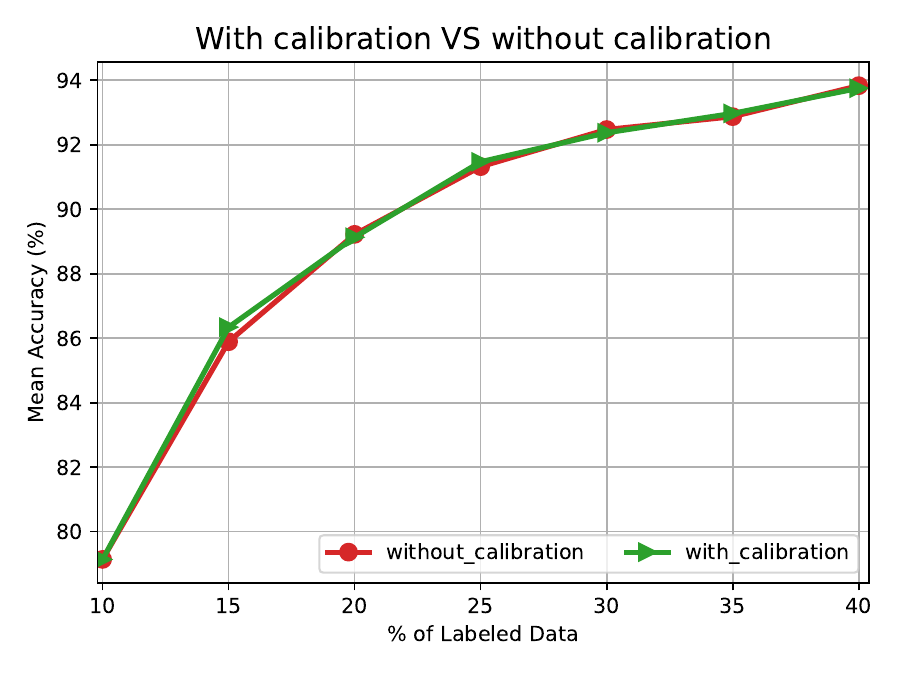}
    \caption{Performance comparison between with and without probability calibration on Cifar10 \cite{krizhevsky2009learning}.
    }
    \label{fig:cali}
\vspace{-0.2cm}
\end{figure}

\noindent
\textbf{Over-fitting Alleviation.}
Here, we compare the generalization of the task model that is trained with the data selected by different AL methods. Specifically, after each AL cycle, we compute the gap between the training and testing accuracy. The smaller the better for this value, since a larger value means the task model to an extent suffers from over-fitting. As can be seen in Table \ref{table:overfit}, our method shows the smallest gap, 
demonstrating that it makes the task model generalize better on testing data. 

\begin{table}[htbp]
\centering
\scriptsize
\caption{Gap (\%) between training and testing accuracy at each AL cycle. \textbf{LB} stands for labeling budget, corresponding to AL cycle. \textbf{M} and \textbf{D} denote method and dataset, respectively.}
\begin{tabular}{|l|ccccccc|}
\hline
\textbf{LB (\%)} & 10 & 15 & 20 & 25 & 30 & 35 & 40 \\
\hline
\diagbox{M}{D} & & & & \textbf{Cifar10} & & &\\
\hline
\textbf{random} & \textbf{19.67} & 14.48 & 12.68 & 10.75 & 9.61 & 8.57 & 8.06  \\


\textbf{core-set \cite{sener2018active}} & 20.81 & 14.59 & 12.73 & 10.88 & 9.95 & 8.92 & 8.31  \\

\textbf{vaal \cite{sinha2019variational}} & 21.07 & 15.01 & 12.24 & 11.08 & 9.66 & 8.57 & 8.08  \\

\textbf{ll4al \cite{yoo2019learning}} & 21.17 & 13.83 & 10.7 & 8.89 & 8.78 & 7.64 & 7.23  \\

\textbf{ours} & 20.41 & \textbf{13.41} & \textbf{10.03} & \textbf{8.4} & \textbf{6.95} & \textbf{6.18} & \textbf{5.67}  \\
\hline

\end{tabular}

\label{table:overfit}
\vspace{-0.4cm}
\end{table}

\noindent
\textbf{Class-wise Performance for Imbalanced Data.}
SVHN \cite{svhn} is a typical imbalanced dataset and our method also yields competitive performance on it as shown in Fig. \ref{fig:classification1} \textbf{Right}. Here, we investigate whether the good performance is dominated by the majority classes. As illustrated in Table \ref{table:class-wise}, while our method works well for the majority classes, it also achieves a good performance on the minority classes. For instance, the highest accuracy of 97.78\% is achieved in class 4 which is a minority class. This demonstrates the advantage of our method on imbalanced data. 

\begin{table}[htbp]
\centering
\footnotesize
\caption{Class-wise classification accuracy (\%) of our method on the testing data of SVHN \cite{svhn}.}
\begin{tabular}{|l|c|c|c|c|c|c|c|c|c|c|}
\hline
\textbf{Class} & 0 & 1 & 2 & 3 & 4  \\
\hline
\textbf{Amount} & 1744 & 5099 & 4149 & 2882 & 2523  \\ 
\hline
\textbf{Accuracy} & 97.53 & 97.49 & 96.99 & 93.79 & 97.78   \\
\hline
\textbf{Class} & 5 & 6 & 7 & 8 & 9 \\
\hline
\textbf{Amount} & 2384 & 1977 & 2019 & 1660 & 1595 \\ 
\hline
\textbf{Accuracy} & 95.47 & 95.45 & 95.99 & 96.02 & 96.36   \\
\hline

\end{tabular}

\label{table:class-wise}
\vspace{-0.1cm}
\end{table}

\begin{table}[htbp]
\centering
\scriptsize
\vspace{-0.2cm}
\caption{Comparison of the active learning methods via the number of the selected samples (out of 2500) of high gradient norm.}
\begin{tabular}{|l|ccccccc|}
\hline
\textbf{LB (\%)} & 10 & 15 & 20 & 25 & 30 & 35 & 40 \\
\hline
\diagbox{M}{D} & & & & \textbf{Cifar10} & & &\\
\hline
\textbf{mc \cite{gal2017deep}} & 462 & 783 & 1162 & 1442 & 1537 & 1600 & 1516  \\

\textbf{core \cite{sener2018active}} & 217 & 230 & 284 & 339 & 329 & 289 & 255  \\

\textbf{vaal \cite{sinha2019variational}} & 268 & 259 & 215 & 231 & 256 & 240 & 256  \\

\textbf{ll4al \cite{yoo2019learning}} & 391 & 893 & 1084 & 1710 & 1698 & 2035 & 1817  \\

\textbf{ours-ent} & \textbf{677} & \textbf{932} & \textbf{1141} & \textbf{1993} & \textbf{2317} & \textbf{2429} & \textbf{2443}  \\

\textbf{ours-est} & \textbf{703} & \textbf{980} & \textbf{1216} & \textbf{1942} & \textbf{2318} & \textbf{2427} & \textbf{2448}  \\
\hline

\end{tabular}

\label{table:influence}
\end{table}

\begin{table}[htbp]
\vspace{-0.3cm}
\centering
\scriptsize
\caption{Gap (\%) between training and testing accuracy at each AL cycle. \textbf{LB} stands for labeling budget, corresponding to AL cycle. \textbf{M} and \textbf{D} denote method and dataset, respectively.}
\begin{tabular}{|l|ccccccc|}
\hline
\textbf{LB (\%)} & 10 & 15 & 20 & 25 & 30 & 35 & 40 \\
\hline
\diagbox{M}{D} & & & & \textbf{Cifar10} & & &\\
\hline
\textbf{random} & \textbf{19.67} & 14.48 & 12.68 & 10.75 & 9.61 & 8.57 & 8.06  \\


\textbf{core-set \cite{sener2018active}} & 20.81 & 14.59 & 12.73 & 10.88 & 9.95 & 8.92 & 8.31  \\

\textbf{vaal \cite{sinha2019variational}} & 21.07 & 15.01 & 12.24 & 11.08 & 9.66 & 8.57 & 8.08  \\

\textbf{ll4al \cite{yoo2019learning}} & 21.17 & 13.83 & 10.7 & 8.89 & 8.78 & 7.64 & 7.23  \\

\textbf{ours-entropy} & 20.35 & 13.05 & \textbf{9.9} & \textbf{7.76} & \textbf{6.79} & 6.29 & \textbf{5.65}  \\

\textbf{ours-estimated} & 20.11 & \textbf{12.74} & 10.04 & 8.06 & 7.05 & \textbf{6.09} & 5.66  \\
\hline

\end{tabular}

\label{table:overfit}
\vspace{-0.2cm}
\end{table}

\begin{table}[htbp]
\centering
\footnotesize
\vspace{-0.2cm}
\caption{Classification accuracy (\%) of ResNet-50 \cite{he2016deep} that is trained on the data selected by ResNet-18. 20,000 samples in the training set of Cifar10 \cite{krizhevsky2009learning} are selected by the different AL methods, respectively. ``mc":  MC-Dropout. ``core": Core-Set.}
\resizebox{\linewidth}{!}{
\begin{tabular}{|l|c|c|c|c|c|c|}
\hline
\textbf{Method} & mc \cite{gal2017deep} & core \cite{sener2018active} & vaal \cite{sinha2019variational} & ll4al \cite{yoo2019learning} & ours-ent & ours-est\\
\hline
\textbf{ResNet-50} & 91.23 & 90.12 & 90.97 & 92.32 & 93.16 & 94.1 \\ 
\hline
\end{tabular}
}
\label{table:verydeep}
\vspace{-0.4cm}
\end{table}

\begin{table}[htbp]
\centering
\vspace{-0.1cm}
\small
\caption{Class-wise classification accuracy (\%) of our-entropy on the testing data of SVHN \cite{svhn}.}
\begin{tabular}{|l|c|c|c|c|c|c|c|c|c|c|}
\hline
\textbf{Class} & 0 & 1 & 2 & 3 & 4  \\

\hline
\textbf{Accuracy} & 97.19 & 96.96 & 97.42 & 93.34 & 97.42   \\
\hline
\textbf{Class} & 5 & 6 & 7 & 8 & 9 \\
\hline

\textbf{Accuracy} & 95.85 & 96.46 & 96.04 & 96.02 & 94.55   \\
\hline

\end{tabular}

\label{table:class-wise}
\vspace{-0.1cm}
\end{table}

\begin{table}[htbp]
\centering
\small
\caption{Class-wise classification accuracy (\%) of our-estimated on the testing data of SVHN \cite{svhn}.}
\begin{tabular}{|l|c|c|c|c|c|c|c|c|c|c|}
\hline
\textbf{Class} & 0 & 1 & 2 & 3 & 4  \\
\hline
\textbf{Accuracy} & 96.85 & 97.51 & 97.06 & 93.82 & 97.23   \\
\hline
\textbf{Class} & 5 & 6 & 7 & 8 & 9 \\
\hline
\textbf{Accuracy} & 95.22 & 96.81 & 95.84 & 96.39 & 95.17   \\
\hline
\end{tabular}

\label{table:class-wise}
\vspace{-0.1cm}
\end{table}

\noindent
\textbf{Probability Calibration.}
As shown in eq. \ref{eq4}, we adopt the KL divergence over output posteriors without conducting probability calibration. As pointed out by Weinberger in \cite{guo2017calibration}, modern neural networks are no longer well-calibrated. But this does not prevent using KL or JS divergence to estimate the distance of output distributions \cite{yu2013kl,kuo2018cost,havasi2018minimal}. 
Therefore, we use the KL divergence \textit{without} probability calibration in the experiments in section \ref{imgclassify} to \ref{cryo-sec}. 

Here, we explore how probability calibration will impact our method when it is used before computing the KL divergence for uncertainty estimation. 
We follow \cite{guo2017calibration} to adopt temperature scaling to calibrate output probability. We conduct this experiment on Cifar10 \cite{krizhevsky2009learning}. To estimate the temperature, we randomly select 5000 samples out of the total 50000 training samples for validation, and the AL task is performed on the remaining 45000 training samples. As illustrated in Fig. \ref{fig:cali}, with and without probability calibration yield quite similar results. 
Therefore, when we compare our method with the AL baselines, we report the results obtained without using probability calibration.  

\noindent
\textbf{Train Deeper Model with Selected Data.}
Here, we investigate whether the data selected by ResNet-18 can effectively train a much deeper model, such as ResNet-50 \cite{he2016deep}. 
As shown in Table \ref{table:verydeep}, 
ResNet-50 achieves the best performance when it is trained on the data selected by our method, further demonstrating that our selected data is more informative and representative.

\begin{table}[htbp]
\centering
\footnotesize
\caption{Classification accuracy (\%) of ResNet-50 \cite{he2016deep} that is trained on the data selected by ResNet-18. 20,000 samples in the training set of Cifar10 \cite{krizhevsky2009learning} are selected by the different AL methods, respectively. ``mc":  MC-Dropout. ``core": Core-Set.}
\begin{tabular}{|l|c|c|c|c|c|}
\hline
\textbf{Method} & mc \cite{gal2017deep} & core \cite{sener2018active} & vaal \cite{sinha2019variational} & ll4al \cite{yoo2019learning} & ours \\
\hline
\textbf{ResNet-50} & 91.23 & 90.12 & 90.97 & 92.32 & 93.33 \\ 
\hline

\end{tabular}

\label{table:verydeep}
\vspace{-0.2cm}
\end{table}

\noindent
\textbf{Time Efficiency.}
Since our method is free of complex auxiliary models or training fashions, it is easy to be deployed and time-efficient. We compare the training time (including data selection) of our method with that of the AL baselines. The results and analysis can be found in the Appendix.

\section{Conclusion}
We propose an effective active learning method exploiting knowledge transfer. We show that data uncertainty is tied to the upper-bound of task loss and present a novel way to select uncertain data for annotation. 
In the experiments, we demonstrate that effective uncertainty estimation makes our method achieve promising results on classical computer vision tasks and cryo-ET challenges. We also analyze the convenience and time efficiency of the proposed method, demonstrating its potentials for various tasks. 

\bibliographystyle{IEEEtran}
\bibliography{IEEE-conference-template-062824}


\end{document}